\documentclass[letterpaper, 10 pt, conference]{ieeeconf}  

\IEEEoverridecommandlockouts                              
\usepackage{graphics} 
\usepackage{subfigure}
\usepackage{amsmath} 
\usepackage{amssymb}  
\usepackage{tikz}
\usepackage{import}
\usepackage{pgfplots}
\pgfplotsset{compat=1.18}
\usepackage{graphicx}
\graphicspath{{./Imgs/}{./Results/}}
\usepackage{todonotes}
\usepackage{url}
\usepackage[utf8]{inputenc}
\newcommand{\Fone}{F1Tenth }
\newcommand{\simreal}{sim-to-real }
\usepackage{array}
\usepackage{booktabs}
\usepackage{tabularx}
\usepackage{multirow}

\newcolumntype{C}{>{\centering\arraybackslash}X}
\newcolumntype{L}{>{\raggedright\arraybackslash}X}

\usetikzlibrary{shapes, shadows, arrows, arrows.meta}
\usetikzlibrary{fit}

\title{\LARGE \bf Bypassing the Simulation-to-reality Gap: Online Reinforcement Learning using a Supervisor}

\author{Benjamin David Evans$^{1}$, Johannes Betz$^{3}$, Hongrui Zheng$^{2}$, Herman A. Engelbrecht$^{1}$,\\ Rahul Mangharam$^{2}$, and Hendrik W. Jordaan$^{1}$
\thanks{$^{1}$ B.D. Evans, H.A. Engelbrecht, and H.W. Jodaan are with the Department of Electrical and Electronic Engineering, Stellenbosch University, South Africa (e-mail: bdevans, hebecht, wjordaan@sun.ac.za)}
\thanks{$^{2}$ H. Zheng, and R. Mangharam are with the Department of Electrical and Systems Engineering, University of Pennsylvania, Philadelphia, USA (e-mail: hongruiz, rahulm@seas.upenn.edu)}
\thanks{$^{3}$ J. Betz, is with the Professorship Autonomous Vehicle Systems, Technical University of Munich, Munich, Germany (e-mail: johannes.betz@tum.de)}
}
\begin{document}

\maketitle
\thispagestyle{empty}
\pagestyle{empty}

\begin{abstract}

Deep reinforcement learning (DRL) is a promising method to learn control policies for robots only from demonstration and experience. 
To cover the whole dynamic behaviour of the robot, DRL training is an active exploration process typically performed in simulation environments. 
Although this simulation training is cheap and fast, applying DRL algorithms to real-world settings is difficult.
If agents are trained until they perform safely in simulation, transferring them to physical systems is difficult due to the sim-to-real gap caused by the difference between the simulation dynamics and the physical robot.
In this paper, we present a method of online training a DRL agent to drive autonomously on a physical vehicle by using a model-based safety supervisor. 
Our solution uses a supervisory system to check if the action selected by the agent is safe or unsafe and ensure that a safe action is always implemented on the vehicle.
With this, we can bypass the sim-to-real problem while training the DRL algorithm safely, quickly, and efficiently. 
We compare our method with conventional learning in simulation and on a physical vehicle.
We provide a variety of real-world experiments where we train online a small-scale vehicle to drive autonomously with no prior simulation training.
The evaluation results show that our method trains agents with improved sample efficiency while never crashing, and the trained agents demonstrate better driving performance than those trained in simulation.
\end{abstract}

\section{Introduction}
\subsection{Motivation}
Deep reinforcement learning (DRL) is a growing, popular method in autonomous system control~\cite{Shwartz}. 
Like humans that learn from experiences over time, DRL algorithms learn control mappings from sensor readings to planning commands using only observations from the environment and reward signals defined by the engineer.
In contrast to humans who learn in the real world, DRL agents are usually trained in simulation.
These simulation environments require accurate sensor and dynamics models to represent the robot and its surrounding environment.
Unfortunately, the accuracy of simulation environments is limited to maintain good computation time, resulting in the \simreal gap when the simulation-trained DRL agent is transferred to a real-world system~\cite{Zhao2020}.

It is desirable to train an agent directly on the robot, thus altogether avoiding the \simreal gap~\cite{zhu2020ingredients}.
An inherent challenge in the online training of DRL algorithms on real-world robots is that DRL algorithms rely on crashing during training, meaning that training on a physical robot is very difficult or nearly impossible~\cite{Dulac-Arnold2021}.
Crashing physical robots is expensive and a safety concern for the surrounding humans~\cite{garcia2015comprehensive}.
Therefore, being able to train DRL agents safely, crash-free onboard physical robots would enable the application of DRL agents to more physical platforms.
Further, it can be expected that bypassing the \simreal gap will lead to improved DRL policies.

\begin{figure}[t]
    \centering
    \includegraphics[width=0.48\textwidth]{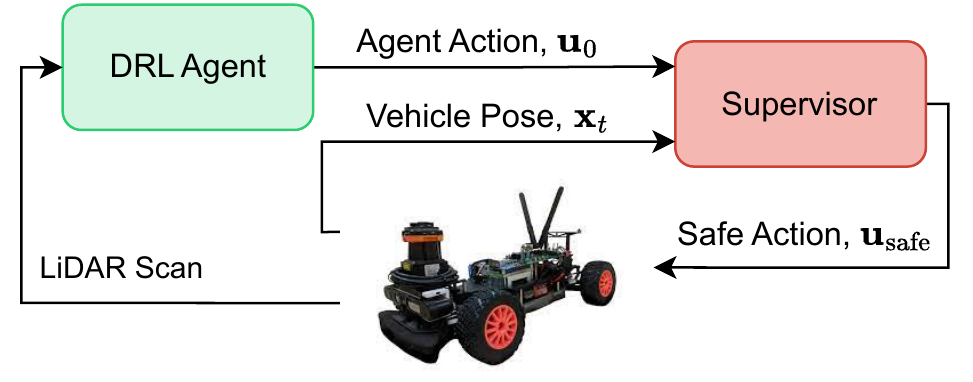}
    \caption{A supervisor ensures the safety of a real-world vehicle during training a DRL agent. The supervisor uses the agent's action and the vehicle's pose to ensure that a safe action is selected.}
    \label{fig:supervisory_architecture}
\end{figure}

\subsection{Contributions}

We address the problem of training DRL agents (with no prior simulation training) on physical vehicles, thus ensuring their safety during the training process.
Figure \ref{fig:supervisory_architecture} shows our approach of using a supervisor to guarantee the vehicle's safety during the DRL agent training.
The supervisory system uses a viability kernel (set of safe states) and vehicle model to check if the DRL agent's action is safe.
If the DRL action is unsafe, a safe action from a pure pursuit controller is implemented. 
After training is completed, the supervisor is removed and the performance of the DRL agent is evaluated.
This work has three main contributions:
\begin{itemize}
    \item We combine a supervisory system with a DRL agent to guarantee crash-free training.

    \item We demonstrate that training an agent with a supervisor results in safe, robust, sample-efficient training of DRL agents in simulation and reality.

    \item We demonstrate agents trained onboard a real-world robot with the supervisor can effectively bypass the \simreal gap by outperforming an agent trained in simulation.

    
\end{itemize}

\section{Related Work}

We discuss works related to DRL for autonomous vehicles, safe DRL training, and online DRL training.

\textbf{DRL for autonomous vehicles:} 
Many variations of DRL-based methods (model-based, model-free) have been implemented to derive control commands for autonomous vehicles from raw sensor inputs. 
The authors of~\cite{Wu2021_2, mirowski2016} used Deep Q-Learning (DQN) to learn steering manoeuvres for autonomous systems and~\cite{Wurman2022-hc, Kuutti2019} used the soft-actor-critic (SAC) algorithm.
Numerous deep learning studies are only evaluated in simulation because they are not practically feasible~\cite{fuchs2021super, jaritz2018end, cai2020high}.
Of the DRL algorithms applied to physical systems, the dominant approach in the literature is to train the agents in simulation before transferring them to real vehicles~\cite{brunnbauer2022latent, Cai2021}. 
Evaluations show that DRL is an effective method of autonomous vehicle control, but the \simreal gap remains a challenge \cite{Chisari2021}.

\textbf{Safe DRL training:} 
In~\cite{Mo2022} and~\cite{wen2020}, a risk-based approach is used to guarantee safety constraints during DRL training. While~\cite{Mo2022} uses a Monte Carlo tree search (MCTS) to reduce unsafe behaviours of the agent while training,~\cite{wen2020}  uses the estimation of trust region constraint to allow large update steps. 
Temporal logic specifications have also been used to enforce safety constraints during training~\cite{cai2022, Alshiekh2018, Li2017}.
A risk-based approach is poorly suited to autonomous vehicles since estimates cannot provide safety guarantees.
Wang et al.~\cite{Wang2022} focuses on ensuring the legal safety of the vehicle by following traffic rules by using a safety layer based on control barrier functions.
Control barrier functions and similar set theory techniques have been used in several safety-critical learning problems~\cite{taylor2020learning}, but have been focused on applications where a safe setting can be assumed~\cite{cheng2019end} or the dynamics can be simplified to linear (affine) equations~\cite{li2018safe}.

\textbf{Online DRL training:}
Training a DRL agent for autonomous driving is difficult since the only input regarding the map is an occupancy grid indicating if a block is open or filled.
Kendal et al.~\cite{Kendall2019} trained a DRL agent on a real-world vehicle using a safety driver (human intervention \cite{Saunders2018}) that decides to intervene if they think the car's position is unsafe. 
Bosello et al.~\cite{bosello2022train} showed that a DRL algorithm on an autonomous vehicle could be trained by simply reversing the vehicle if it was near to crashing.
These approaches demonstrate that online training for autonomous robots is a viable idea but is limited by a simplistic safety system.
Musau et al.~\cite{musau2022using} used formal reachability theory to enable online training on a small-scale vehicle.
Their method estimates future states of the vehicle in real-time resulting in it being computationally intensive and poorly suited to onboard hardware with limited computation.

In summary, safe online DRL training is a growing field that requires further investigation to explore how DRL agents can be trained onboard real-world robots while guaranteeing safety at the same time.
Viable approaches should use the track occupancy grid to determine when the vehicle is on the edge of safety and only then intervene.

\section{Methodology}

\subsection{\Fone Platform}

\Fone racing cars are 1/10th the size of real F1 vehicles and are used as a test-bed for autonomous algorithms~\cite{Betz2022_RacingSurvey}. 
The platform focuses on safe algorithms that run autonomously onboard the vehicle.
The cars are equipped with a LiDAR scanner for sensing the environment, an NVIDIA Jetson NX as the main computation platform, a variable electronic speed controller (VESC) and drive motor to move the vehicle forwards, and a servo motor to steer the front wheels.
The vehicle uses the ROS2 middleware for the sensors, software components and control signals to communicate with each other.

\textbf{Problem:} 
We approach the problem of training a DRL agent to drive a \Fone vehicle autonomously around a provided race track.
The task of planning is to use the onboard sensor measurements, LiDAR scan and odometry (estimated using a particle filter~\cite{walsh17}) to calculate an optimal steering angle $\delta$ and velocity $v$ that results in the vehicle driving around the track.
Training a DRL agent means randomly initializing a policy (neural network), then using the policy to collect experience, and using the collected samples to adjust the policy parameters until the agent can drive around the track.

\textbf{Vehicle Model:} The vehicle is a controlled discrete-time system such that $\mathbf{x}_{k+1} = f(\mathbf{x}_k, \mathbf{u}).$ 
The vehicle state at the current timestep $\mathbf{x}_k$, comprises the vehicle location in the $x$ and $y$ directions and the vehicle orientation, such that $\mathbf{x}_k=[X, Y, \theta]$.
The vehicle control $\mathbf{u}$ consists of a steering angle $\delta$, and velocity $v$, such that $\mathbf{u} = [\delta, v]$.
In our experiments, the vehicle speed is kept constant.
State updates are performed using the single-track vehicle model~\cite{althoff2017commonroad}.
The additional state variables in the single-track model (e.g. slip angle) are set to 0 before each update.

\subsection{Supervisory Safety System}
In contrast to solutions that train DRL agents in simulation and transfer them to physical vehicles~\cite{brunnbauer2022latent}, we present a safety supervisor that enables the training of DRL agents onboard the physical vehicle.
We present the training architecture followed by a description of the supervisor operation.


\begin{figure}[b]
    \centering
    \includegraphics[width=0.48\textwidth]{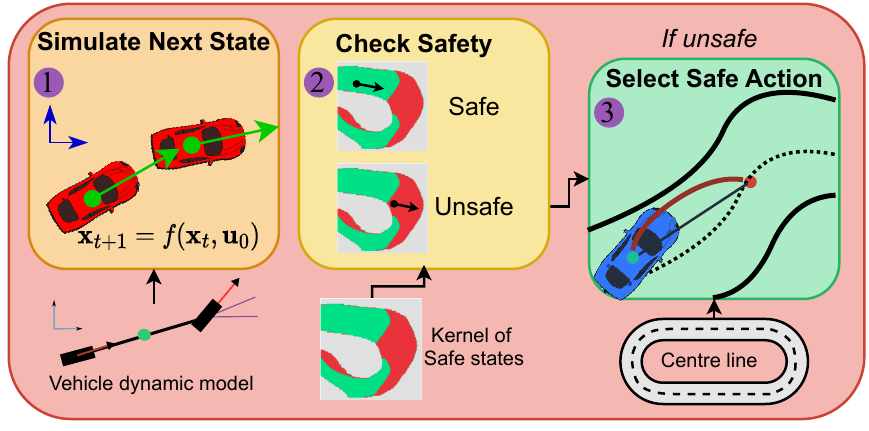}
    \caption{The supervisor ensures safety by simulating the next state, checking if the resulting state is safe and if unsafe, then selecting a safe action.}
    \label{fig:interal_supervisor}
\end{figure}

\textbf{Supervisory Training Architecture:} 
The training architecture, shown in Figure \ref{fig:supervisory_architecture}, uses the supervisor to monitor the agent and ensure that only safe actions are implemented on the vehicle.
Safe actions are those that do not lead the vehicle to crash into the boundary and are recursively feasible, i.e. after taking a safe action; another safe action will definitely exist.
When training is complete, the agent is tested by removing the supervisor to demonstrate that the agent has learned to drive safely around the track.

\textbf{Supervisor Operation:} 
Figure \ref{fig:interal_supervisor} shows how the supervisor fulfils its role of ensuring that only safe actions are implemented on the vehicle through a three-step process of (1) using the current state and action to calculate the next state, (2) checking if the next state is safe, and (3) if unsafe, selecting a safe action.

The supervisor checks if an action $\textbf{u}_0$ is safe by using the current vehicle pose $\textbf{x}_t$ and a dynamics model to simulate the next state where the vehicle will be after the planning timestep.
A kernel of all the possible states (positions and orientation) of the vehicle on the map is divided into the subsets of safe states $\mathcal{X}_\text{safe}$ and unsafe states through a process described in Section \ref{subsec:safe_set_generation}.
The next state is evaluated for safety by checking if it is in the subset of safe states.
If the next state is safe, then the agent action can be implemented; otherwise, a safe action  must be selected.

Selecting a safe action $\textbf{u}_\text{safe}$ is done using a pure pursuit controller~\cite{coulter1992implementation} that uses the single-track model of the car to calculate a steering angle that follows the centerline of the race track~\cite{Heilmeier2019}.
Using the pure pursuit controller in this way ensures that the vehicle will always move towards the center of the track where it is safe, away from the potential danger of the track boundaries.

\subsection{Supervisory Reinforcement Learning} \label{subsec:super_rl}

Reinforcement learning problems are modelled as Markov Decision Processes (MDPs) having a state space, action space, reward signal and transition probability.
During training, the agent, consisting of a neural network, receives a state and selects an action that is implemented and a new state and reward are returned to the agent.
The agent's experience of states, actions, next states and rewards is stored in a buffer and used to update the neural network parameters to select actions that maximise the reward signal.
We use the twin-delayed-deep-deterministic-policy-gradient (TD3) algorithm \cite{fujimoto2018addressing} to train our agents to select continuous control actions.

The DRL agent uses neural networks with two fully connected hidden layers of 100 neurons each and the \textit{ReLu} activation function.
The agent's state vector (input) consists of 20 evenly sliced beams from the LiDAR scan scaled from the LiDAR beam range of 10~m to the range [0, 1].
The output is the steering angle, scaled from [-1, 1] (realized using the \textit{tanh} activation function) to the steering angle range of {0.4~rad}.
The planners operate at a frequency of 10~Hz.

\textbf{Training Reformulation:} 
In conventional RL, an episode is an ordered set (or trajectory) of state, action, reward, and done tuples, from an initial state to a terminal state (crashing or completing a lap).
Using the supervisor, the agent never crashes and always completes laps.
Additionally, if the supervisor intervenes, a different action is implemented on the vehicle to that which the agent selected, breaking the link between state, action and next state.

\begin{figure}[h]
    \centering
    \includegraphics[width=0.48\textwidth]{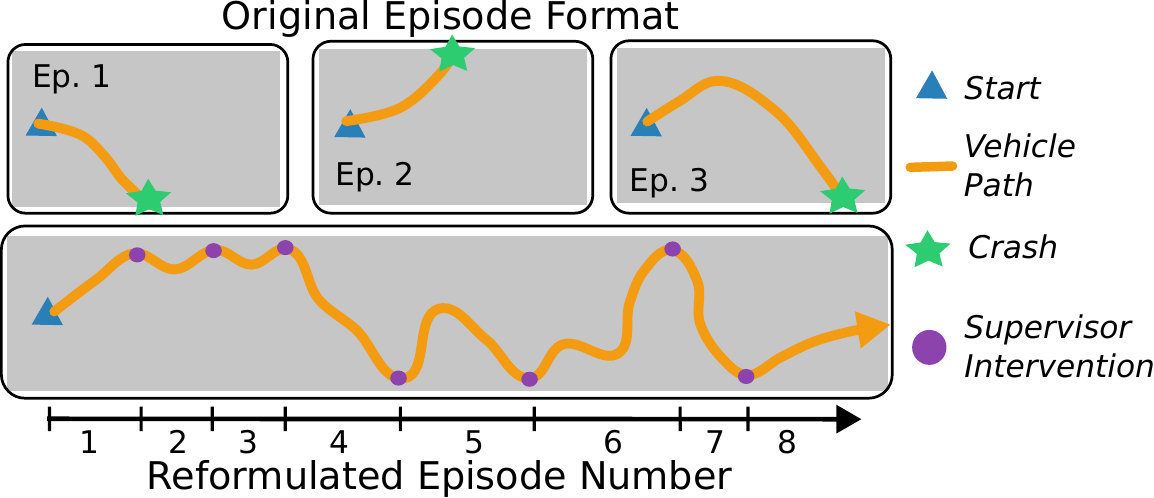}
    \caption{Example vehicle paths comparing the original episode format of crashing and resetting against the reformulated episodes of the supervisor intervening.}
    \label{fig:episode_configuration}
\end{figure}

Using the supervisor, we define an episode to run from the initial state until the supervisor intervenes.
When the supervisor intervenes, it is recorded as a terminal state, and the supervisor gives the agent a penalty of -1.
This penalty for unsafe actions is the only reward used by our method.
Figure \ref{fig:episode_configuration} shows how the definition of an episode has been changed from conventionally requiring many episodes with resetting the vehicle to shorter episodes running while the supervisor does not intervene.

\subsection{Safety Kernel Generation} \label{subsec:safe_set_generation}

The supervisor uses a list (or kernel) of recursively safe states to ensure vehicle safety.
As previously mentioned, safe states are defined as states that do not lead to the vehicle crashing into the boundary and are recursively feasible, meaning that every safe state has an action that leads to another safe state.
The formulation and generation of the kernel of safe states are based on the work by Liniger et al.~\cite{liniger2017real}.

The state space $\mathcal{X}$ is discretized into a countable number of states $\mathcal{X}_h$.
The track map is split into a finite number of blocks by gridding the map with a uniform grid with a resolution of 40 blocks per meter.
The orientation angle $\theta$ is split into 41 even angle segments.
The control space $U$, consisting of the steering angle range, is split into 9 evenly spaced control modes.
The discrete states are used to formulate the dynamics as a difference inclusion where the next state $\mathbf{x}_{k+1}$ is in the set of possible next states, $\mathbf{x}_{k+1} \in F(\mathbf{x}_k)$.
For a given state, the set of all possible next states is written as $F(\mathbf{x}_k) = \{f(\mathbf{x}_k, \delta)~|~\delta \in [-\delta_\text{max}, \delta_\text{max}] \}$.

The kernel of safe states $\mathcal{X}_\text{safe}$ is a subset of the discrete state space $\mathcal{X}_h$, for which there exists a safe action. 
The kernel is calculated using the recursive viability kernel algorithm, 
\begin{equation}
    \begin{split}
    K^0 & = K_\text{track} \\    
        K^{i+1} & = \{\mathbf{x}_h \in K^i~|~\forall~ F(\mathbf{x}_h) \cap K^i \neq \varnothing  \}. \\
    \end{split}
    \label{eqn:viab_algo}
\end{equation}

\begin{figure}[h]
    \centering
    \includegraphics[width=0.45\textwidth]{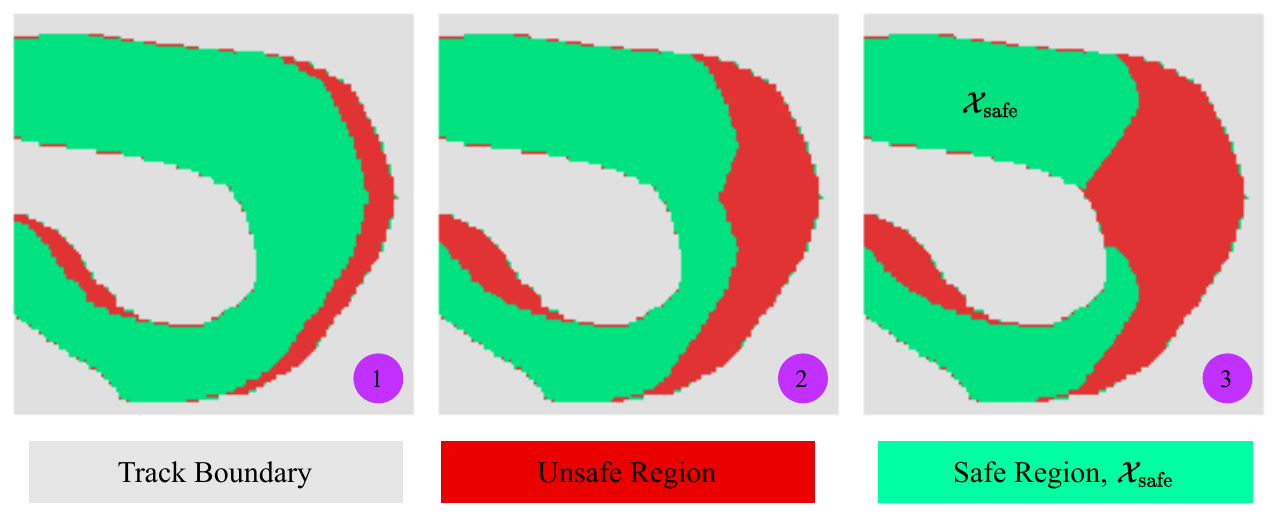}
    \caption{Three stages of kernel growth for a corner on a race track with the vehicle pointing towards the right. Note that the kernel shape (and safe region) depends on the vehicle orientation.}
    \label{fig:growing_kernel}
\end{figure}

The viability kernel algorithm in Equation \ref{eqn:viab_algo} generates a set of states for which there recursively exists an action that causes the vehicle to remain within the kernel.
The algorithm's safe set is initialized ($K^0$) to all the states on the drive-able area of the race track being safe, $K_\text{track}$.
The algorithm then recursively generates smaller safe sets by looping through each safe state from the previous iteration and including only states for which there exists an action that leads to another safe state.
Formally, this process is defined as selecting states for which the intersection of the next states and the kernel is not equal to the empty set. 
This process results in a 3-dimensional kernel of recursively feasible safe states, $\mathcal{X}_\text{safe}$.
Figure \ref{fig:growing_kernel} shows how the kernel grows inwards from the track boundaries until all remaining states are safe.
While the kernel shape depends on the vehicle orientation, Figure \ref{fig:growing_kernel} visualises the kernel with the vehicle orientation pointing towards the right.

\section{Evaluation}

We compare our method of training using a supervisor against the baseline of conventional learning.
Firstly, we compare the difference in training and performance in simulation, then we study online learning onboard a physical vehicle and finally we compare our method against an agent trained in simulation on a real-world hardware platform.

\textbf{Baseline:}
The baseline agent is trained with conventional reinforcement learning where it crashes and is reset to a starting position.
The baseline agent uses a reward signal with a punishment of -1 for crashing, a reward of 1 for completing a lap and a shaped reward relative to the centerline progress (same as \cite{brunnbauer2022latent}).
The shaped reward is calculated as, $r_t = (p_{t} - p_{t-1})/p_\text{total}$, where $p_t$ is the centerline progress at timestep $t$, scaled according to the total centerline length $p_\text{total}$.
For completed laps, the baseline agent receives a reward of 2, where 1 is the sum of intermediate progress rewards and 1 is for lap completion.

\subsection{Simulation Tests}

Conventional and online learning are compared in the open-source F1Tenth simulator~\cite{o2020f1tenth} on four scaled F1 race tracks.
The experiments are repeated three times using different random seeds and the average used, and for each repetition, 20 test laps are completed.
The simulation results use a constant speed of 2 m/s.
Table \ref{tab:map_description} shows the shape of the AUT, MCO, GBR, and ESP tracks (from \cite{bosello2022train}), with the mean times used for the lap time normalisation.

\setlength{\textfloatsep}{0.01cm} 
\begin{table}[h]
    \centering
    \renewcommand{\arraystretch}{1.2}
    \begin{tabular}{@{} m{2cm} m{1.5cm}<{\centering} @{} m{1.5cm}<{\centering} @{} m{1.5cm}<{\centering} @{} m{1.5cm}<{\centering} @{}}
    \toprule
    Track & AUT & ESP & GBR & MCO \\
    \midrule
     Image    & \includegraphics[width=1.2cm]{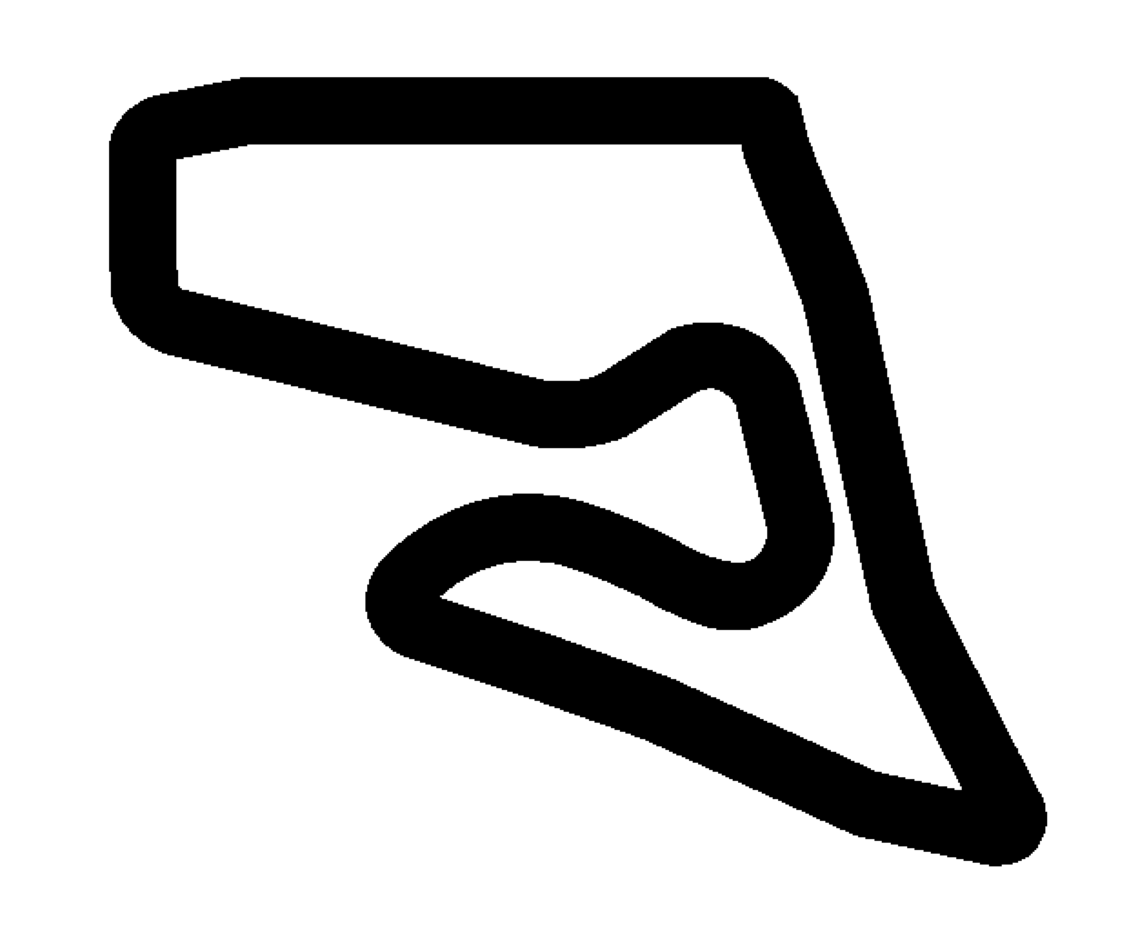}  & \includegraphics[width=1.4cm]{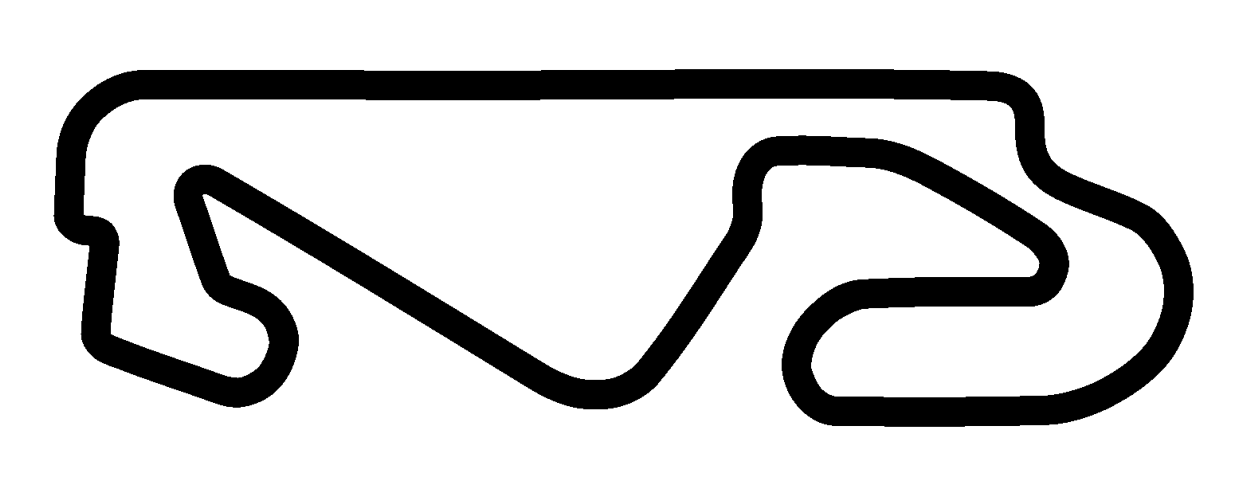}   & \includegraphics[width=1.4cm]{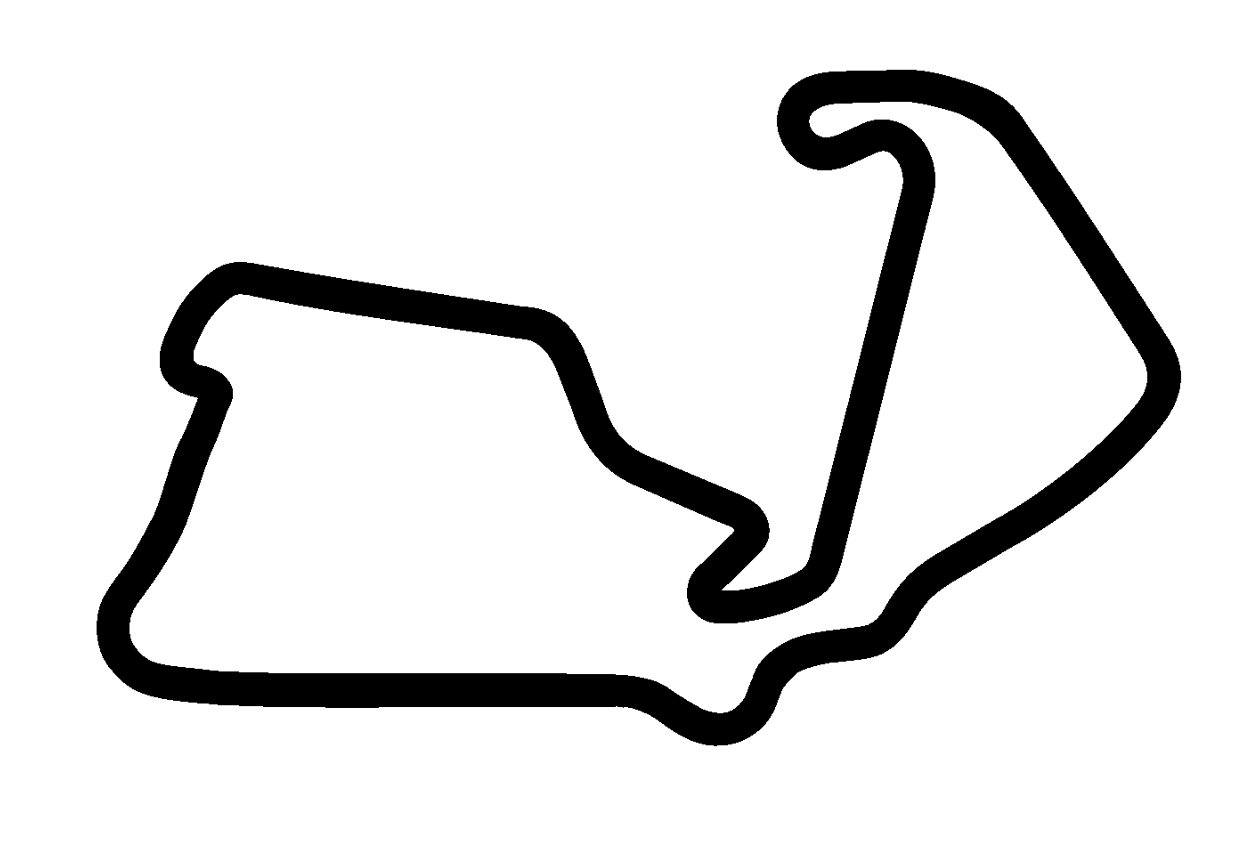}  & \includegraphics[width=1.2cm]{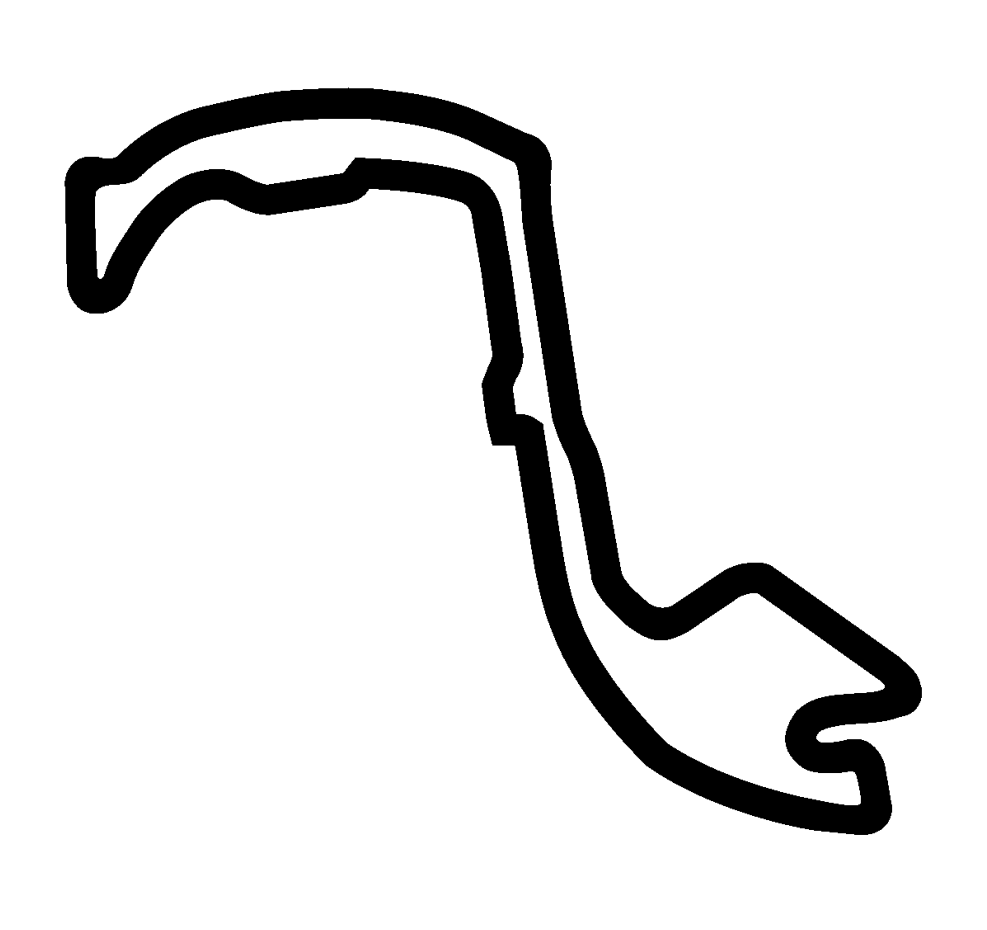}  \\
     Mean lap time (s) & 46.7 & 116.4 & 100.2 &  86.6 \\
     \bottomrule
    \end{tabular}
    \caption{Track images, and mean lap times for the AUT, ESP, GBR and MCO maps.}
    \label{tab:map_description}
\end{table}

\begin{figure}[h]
    \centering
    \includegraphics[width=0.48\textwidth]{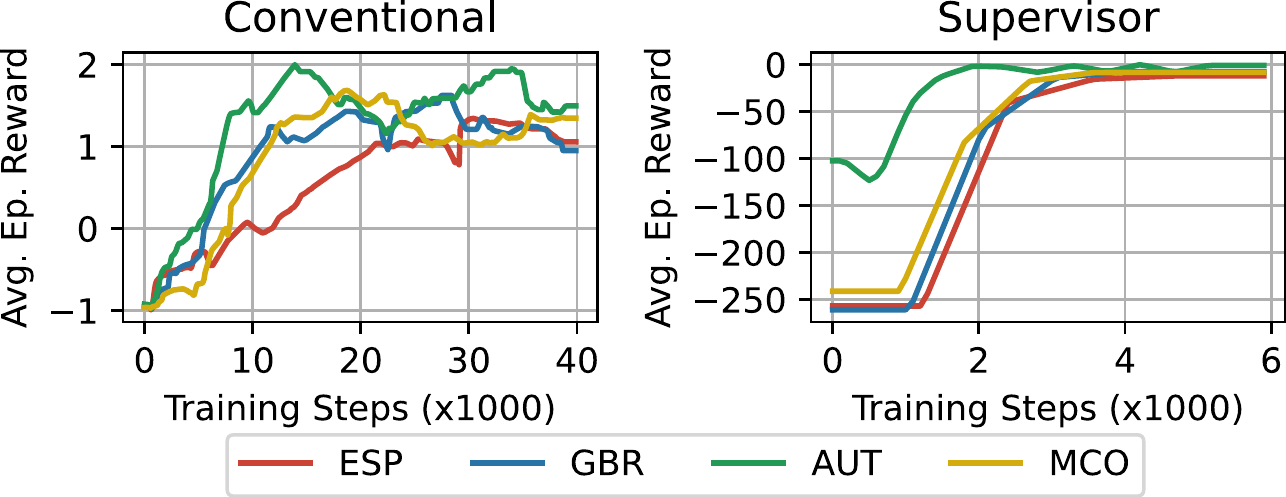}
    \caption{Rewards earned during training of the baseline (left) and online (right) agents on the AUT, MCO, GBR and ESP maps in simulation.}
    \label{fig:simulted_training}
\end{figure}

\textbf{Training Comparison:}
Figure \ref{fig:simulted_training} shows the average episode rewards earned during the training of the baseline agents for 40,000 steps (left) and the online agents for 6,000 steps (right).
The conventional agents start with earning a reward near -1, indicating they crash quickly.
After around 20k steps, the average reward converges to between 1 and 2 with the agents trained on the AUT track receiving the most reward and the agents on the MCO track receiving the least reward.
The online agents start earning large negative rewards of around -100 or the AUT track and -250 for the other tracks, indicating that the supervisor intervenes a lot at the beginning of training.
After around 4k training steps the agents all receive near to 0 reward indicating that the supervisor seldom intervenes.

Therefore, the simulated training comparison demonstrates that online training is more sample efficient than conventional training requiring only 6k training steps.

\begin{figure}[h]
    \centering
    \includegraphics[width=0.48\textwidth]{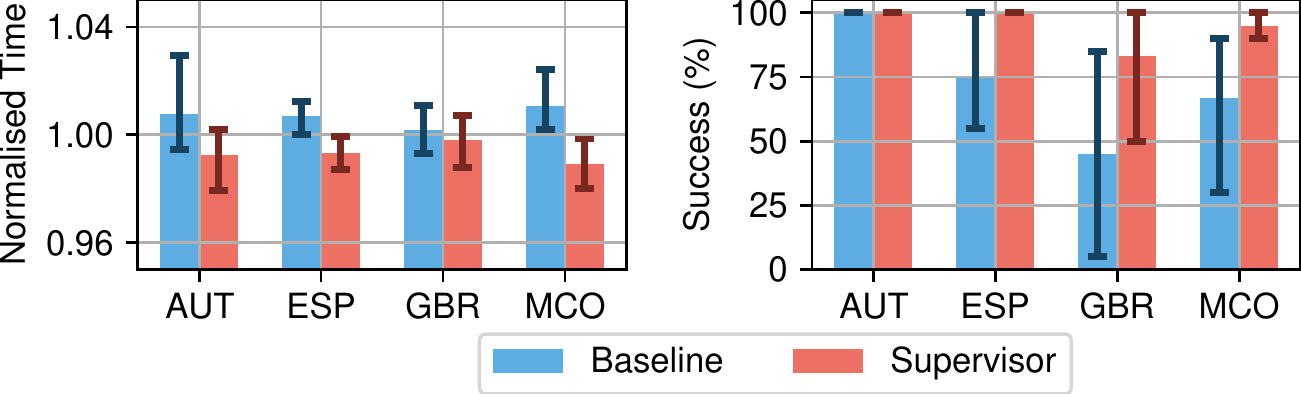}
    \caption{Lap times and total curvature of the baseline and online agents compared to a pure pursuit planner.}
    \label{fig:simulation_performance_results}
\end{figure}

\textbf{Performance Comparison:}
Figure \ref{fig:simulation_performance_results} shows bar plots of the normalised lap times and success rates of the baseline and online agents.
The error bars represent the minimum and maximum values from the three repetitions.
The lap times are normalised by dividing them by the mean times shown in Table \ref{tab:map_description}.
The success rate is the percentage of the test laps that were completed without the vehicle colliding.

\begin{figure*}[t]
    \centering
    \begin{minipage}{0.34\textwidth}
        \centering
        \vspace{1mm}
        \rotatebox{0}{\input{Results/e_2_train_lap_train_0.tex}}
        \vspace{3mm}
        \caption{Training lap (starting at the red cross) of the safety agent (blue line) with the locations where the agent stopped to train (yellow dots).}
        \label{fig:training_lap_e2}
    \end{minipage}
    \hfill
    \begin{minipage}{0.62\textwidth}
        \centering
        \vspace{1mm}
        \scalebox{0.88}{\input{Results/steering_graph.tex}}
        \caption{Comparison of steering actions selected by the agent (green) and those implemented by the safety system (blue) during training of the safety agent in simulation in environment 1.}
        \label{fig:steering_actions}
    \end{minipage}
\end{figure*}

In Figure \ref{fig:simulation_performance_results}, the agents trained with the supervisor achieve lower normalised lap times on all of the maps.
The agents trained with the supervisor have a higher percentage success rate.
For example, on the MCO track, the agent trained with the supervisor achieves a 95\% average completion rate, while the conventionally trained agent achieves only 70\%.
The simulation results indicate that training with a supervisor results in faster lap times with higher success rates.

\subsection{Real-world Tests}

The supervisor is used to train a DRL agent, with no a priori knowledge or training, onboard a vehicle to drive around a track autonomously.
The vehicle is trained by completing two laps in environment 1 (around 800 steps), driving at a constant speed of 2 m/s.
Figure \ref{fig:training_lap_e2} shows the first training lap of a randomly initialised agent being trained online using the supervisor.
The trajectory shows the agent's squiggles as it veers to one side and then to the other.
Due to the computation burden of training the agent, the vehicle collects 20 steps of experience and then stops to train before continuing to collect more data (shown by yellow dots).
The entire training process of driving two laps while stopping to train takes around 10 minutes.


\textbf{Supervisor Effect:} 
A graph comparing the steering angles selected by the agent (green) and the safe actions implemented by the supervisor (blue) is shown in Figure \ref{fig:steering_actions}. 
At the beginning of the training, the agent rarely selects safe actions.
As the training progresses, the agent selects more safe actions, and towards the end, the agent rarely selects an unsafe action.
This result shows the supervisor's behaviour in preventing the agent from taking unsafe actions during the initial stage of training, and how, as the agent is trained, it learns to select safe actions without requiring the supervisor.

\textbf{Online Training Rewards:}
We investigate the rewards earned by the agent during training with the supervisory system.
We use the sum of the reward achieved every 20 steps (the interval of data collection between the agent stopping to train) as the metric to measure the online training performance.
The worst reward is -20 if the supervisor intervenes at every step, and the maximum reward is 0 if the supervisor never intervenes.

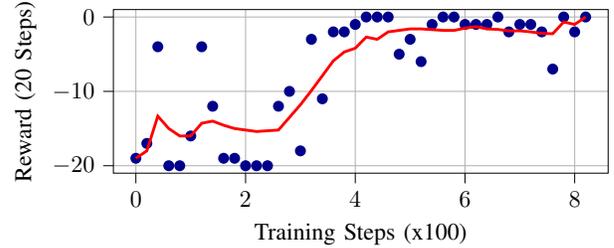
\begin{figure}[h]
    \centering
    \scalebox{0.9}{
\begin{tikzpicture}

\definecolor{darkblue}{RGB}{0,0,139}
\definecolor{darkgray176}{RGB}{176,176,176}

\begin{axis}[
clip mode=individual,
height=4cm,
minor xtick={},
minor ytick={},
tick align=outside,
tick pos=left,
width=0.5\textwidth,
x grid style={darkgray176},
xlabel={Training Steps (x100)},
xmajorgrids,
xmin=-0.41, xmax=8.61,
xtick style={color=black},
xtick={-2,0,2,4,6,8,10},
y grid style={darkgray176},
ylabel={Reward (20 Steps)},
ymajorgrids,
ymin=-21, ymax=1,
ytick style={color=black},
ytick={-30,-20,-10,0,10}
]
\addplot [semithick, darkblue, mark=*, mark size=2, mark options={solid}, only marks]
table {%
0 -19
0.2 -17
0.4 -4
0.6 -20
0.8 -20
1 -16
1.2 -4
1.4 -12
1.6 -19
1.8 -19
2 -20
2.2 -20
2.4 -20
2.6 -12
2.8 -10
3 -18
3.2 -3
3.4 -11
3.6 -2
3.8 -2
4 -1
4.2 0
4.4 0
4.6 0
4.8 -5
5 -3
5.2 -6
5.4 -1
5.6 0
5.8 0
6 -1
6.2 -1
6.4 -1
6.6 0
6.8 -2
7 -1
7.2 -1
7.4 -2
7.6 -7
7.8 0
8 -2
8.2 0
};
\addplot [very thick, red]
table {%
0 -19
0.2 -18
0.4 -13.3333333333333
0.6 -15
0.8 -16
1 -16
1.2 -14.2857142857143
1.4 -14
1.6 -14.5555555555556
1.8 -15
2 -15.2
2.2 -15.4
2.4 -15.3
2.6 -15.2
2.8 -13.5
3 -11.8
3.2 -9.9
3.4 -7.9
3.6 -5.9
3.8 -4.7
4 -4.2
4.2 -2.7
4.4 -3
4.6 -2
4.8 -1.8
5 -1.6
5.2 -1.6
5.4 -1.7
5.6 -1.8
5.8 -1.8
6 -1.5
6.2 -1.3
6.4 -1.6
6.6 -1.66666666666667
6.8 -1.875
7 -1.85714285714286
7.2 -2
7.4 -2.2
7.6 -2.25
7.8 -0.666666666666667
8 -1
8.2 0
};
\end{axis}

\end{tikzpicture}}
    \caption{Training rewards per 20 steps for safety agent trained on the physical vehicle (blue dots) with moving average (red).}
    \label{fig:safety_training_graph}
\end{figure}

Figure \ref{fig:safety_training_graph} shows a graph of the sum of rewards achieved every 20 steps by the agent trained online the physical vehicle.
The graph shows that in the beginning, the agent receives low rewards; as time progresses, the agent receives higher rewards.
After around only 400 steps, the agent displays a significant improvement, which corresponds to the graph of safe steering actions in Figure \ref{fig:steering_actions} showing that the agent requires less intervention between 300-400 training steps.
This result shows that our method of using a supervisor is effective for training a DRL agent onboard a real-world vehicle in only 800 training steps.






\setlength{\textfloatsep}{0.01cm} 
\begingroup
\begin{table*}[]
    \centering
    \renewcommand{\arraystretch}{1.4}
    \begin{tabular}{w{l}{4cm} cccccccccc}
    \toprule
                & \multicolumn{4}{c}{\textit{Environment 1}}                                      &  &   & \multicolumn{4}{c}{\textit{Environment 2}}        \\                  
        \cmidrule{2-5} 
        \cmidrule{8-11}
      & \multicolumn{2}{c}{\textbf{Simulation}} & \multicolumn{2}{c}{\textbf{Reality}} & & &   \multicolumn{2}{c}{\textbf{Simulation}} & \multicolumn{2}{c}{\textbf{Reality}} \\
                                         & Baseline    & Supervisor   & Baseline             & Supervisor       &    &      &      Baseline    & Supervisor & Baseline            & Supervisor            \\ \hline
  \multicolumn{1}{l|}{Distance Driven in m}      &    65.0          & \textbf{59.8}     &    65.68     &    \multicolumn{1}{c}{\textbf{61.6}} & \multicolumn{1}{c||}{} & &   17.0    &   \textbf{15.6}    &        18.8       &    \textbf{17.3}        \\
 \multicolumn{1}{l|}{Lap-time in s }      &  32.8       & \textbf{31.1}    &    35.5      &    \multicolumn{1}{c}{ \textbf{32.5}}   & \multicolumn{1}{c||}{} & &  12.9        &   \textbf{11.1}  &       12.8        &    \textbf{10.1}            \\
  \multicolumn{1}{l|}{Mean Steering Angle in rad } &      0.30        &  \textbf{0.03} &    0.22      &     \multicolumn{1}{c}{ \textbf{0.06 }}  & \multicolumn{1}{c||}{} & &    0.36        & \textbf{0.11}      &       0.31         &    \textbf{0.09}          \\
  \multicolumn{1}{l|}{Total Curvature in m$^{-1}$}&     274.6    &    \textbf{34.2}  &    207.5       &     \multicolumn{1}{c}{\textbf{86.3}}   & \multicolumn{1}{c||}{} & &   119.1       & \textbf{44.9  }    &       86.6            &  \textbf{49.3 }            \\
        \bottomrule
    \end{tabular}
    \caption{Quantitative comparison of online and offline trained DRL agents in two different environments.}
    \label{tab:safety_tests}
\end{table*}
\endgroup

\textbf{Quantitative Analysis:} 
We compare the performance of a DRL agent trained online a physical vehicle with the safety system against a baseline agent, trained offline in a simulator and then transferred to the vehicle.
The baseline agent is trained in simulation on the environment 1 map for 30,000 steps.

\begin{figure}[h]
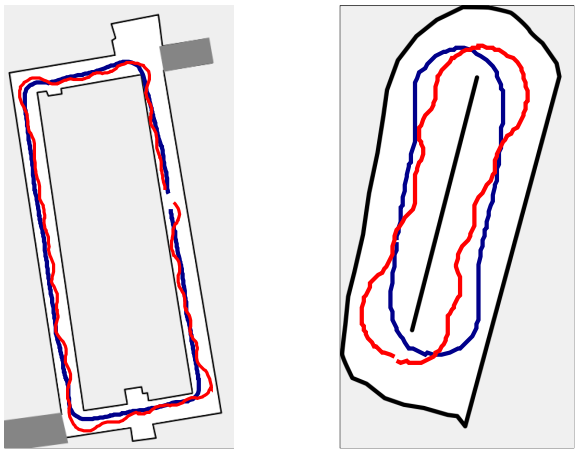

    \centering
    \begin{minipage}{0.48\linewidth}
        \centering
        \rotatebox{90}{\scalebox{1}{\input{Results/e_4_test4_lap_0.tex}}}
    \end{minipage}
    \hfill
    \begin{minipage}{0.48\linewidth}
        \centering
        \rotatebox{90}{{\input{Results/ex4_on_lobby_test_lap_2.tex}}}
    \end{minipage}
    \caption{Comparing the test trajectories of the baseline (red) and online (blue) agents on environment 1 (left) and environment 2 (right)}
    \label{fig:trajectory_comparison}
\end{figure}

Table \ref{tab:safety_tests} presents the quantitative results of the offline (baseline), and online (supervisor) trained DRL agents in two different environments with the metrics of distance travelled, lap time, absolute mean steering and curvature. 
The agent trained with the supervisor generally leads to a lower mean steering angle and lower total curvature of the trajectories, resulting in lower distance travelled and lower corresponding lap times than the baseline agents.
For example, on the physical vehicle driving in environment 1 (shown in Figure \ref{fig:trajectory_comparison}), the baseline agent travelled 65.0 m, while the supervisory agent travelled only 59.8 m, which is 5.2 m shorter. 
The average steering angle for the supervisory agent was 0.03 radians, compared to the mean steering angle for the baseline of 0.3 radians.
The baseline total curvature was significantly more (207.5) than the supervisory agent's (86.3).

This result demonstrates that training agents with the supervisor outperforms conventionally trained DRL agents on real-world vehicles with smoother steering actions.
Although the supervisory agent also performs worse in reality compared to simulation, training the DRL agent on the real car shows a definite improvement in the performance of the physical vehicle compared to the baseline.

\textbf{Qualitative Analysis:} 
Figure \ref{fig:trajectory_comparison} shows the real-world trajectories for the baseline agent trained in simulation (red) and our method trained onboard the vehicle (blue) in Figure \ref{fig:trajectory_comparison}.
The left image shows the test trajectories in environment 1 (the training track).
The baseline selects a squiggly, un-smooth path, regularly coming close to the track boundaries and almost crashing.
In contrast, the DRL agent trained onboard the vehicle using the supervisor has a much smoother trajectory, driving in a line through the straights and smoothly turning the corners.


\textbf{Robustness:} 
A crucial aspect of DRL agents is their ability to learn general policies that can be transferred to other environments. 
Both the agents trained in simulation and on the physical vehicle are tested on the track they were trained on (environment 1) and a different test track (environment 2, right in Figure \ref{fig:trajectory_comparison}).
Due to the reduced size of the track, the speed was reduced to 1.5 m/s.
The first observation is that both the baseline and supervisory agents can complete laps on a different track to the one that they were trained on, highlighting the advantage of the flexibility and adequate generalization of DRL agents.
The right image in Figure \ref{fig:trajectory_comparison} shows that the trajectories followed in environment 2 display a similar pattern to that of environment 1.
The supervisory agent takes a smoother path and swerves less than the baseline.
This outcome is reinforced by the quantitative results in Table \ref{tab:safety_tests}, which show that the agent trained with the supervisor achieves a shorter lap time (1.5~s different), with a lower mean steering angle (0.09 versus 0.31) and less total curvature (49.3 versus 86.6) than the baseline planner.
Therefore, we conclude that training a DRL agent onboard with a supervisor results in more general behaviour, as demonstrated by improving performance on a different track.

\section{Conclusion}

This paper presented a supervisory safety system capable of training a DRL agent for autonomous driving online on the vehicle car. 
The supervisor ensures vehicle safety by checking if the DRL agent's action is safe or unsafe, using a pure pursuit planner to select a safe action.
We did two evaluations to prove the algorithm's robustness, once in simulation and once on a physical real-world vehicle. The evaluation in simulation demonstrated that using the supervisor to train agents results in lower lap times and higher success rates while requiring fewer training steps. The real-world test demonstrated that the supervisory system is effective for safely training a randomly initialized agent onboard a physical vehicle. The autonomous vehicle demonstrated a safely driven path in the given environment.
The results showed that the agent trained online with the supervisor performed better than the agent trained purely in simulation by driving a shorter path around the track, and selecting a smoother path than the baseline without requiring additional measures (such as reward hacking in \cite{bosello2022train} or action regularisation in \cite{brunnbauer2022latent}).
The agent trained with the supervisor could transfer to an environment unseen during training where it outperformed the conventionally trained agent.
These results demonstrate that our method effectively bypasses the \simreal gap by training agents onboard real-world vehicles.

Future work should address expanding safe learning onboard physical robots to other physical platforms such as UAV control and high-speed autonomous racing. 
This task requires extending the principle of formulating a supervisor and a safety policy to operate the robotic system at its performance limits and include more control actions, e.g. velocity, acceleration, and various vehicle dynamics parameters.



\typeout{}

\end{document}